\documentclass[11pt]{article}
\usepackage{acl2016}
\usepackage{url}
\usepackage{times}
\usepackage{latexsym}
\usepackage{amsmath, amssymb}
\usepackage{graphicx,caption,subcaption}
\usepackage{booktabs}
\usepackage[hang,flushmargin]{footmisc}
\usepackage{soul}  
\usepackage[draft]{pgf}  
\usepackage{color}
\usepackage{xcolor}
\definecolor{darkred}{rgb}{0.5,0,0}
\definecolor{darkblue}{rgb}{0,0,0.5}
\usepackage{hyperref}
\usepackage{array}

\aclfinalcopy 

\newcommand\sssection[1]{\vspace{0.5em}\noindent \textbf{#1} \hspace{0.4em}}

\title{Neural Language Correction with Character-Based Attention}

\author{Ziang Xie,  Anand Avati, Naveen Arivazhagan, Dan Jurafsky, Andrew Y. Ng\\
        Computer Science Department, Stanford University\\
        {\tt \{zxie,avati,naveen67,ang\}@cs.stanford.edu, jurafsky@stanford.edu}}

\date{}

\begin{document}

\maketitle

\begin{abstract}

Natural language correction has the potential
to help language learners improve their writing skills.
While approaches with separate classifiers
for different error types
have high precision, they do not flexibly handle errors
such as redundancy or non-idiomatic phrasing.
On the other hand,
word and phrase-based machine translation methods
are not designed to cope with orthographic errors,
and have recently been outpaced by neural models.
Motivated by these issues, we present a neural
network-based approach to language correction.
The core component of our method is an encoder-decoder
recurrent neural network with
an attention mechanism.
By operating at the character level,
the network avoids the problem of out-of-vocabulary words.
We illustrate the flexibility of our approach on dataset
of noisy, user-generated text collected from an English learner forum.
When combined with a language model,
our method achieves a state-of-the-art $F_{0.5}$-score on the CoNLL 2014 Shared Task. 
We further illustrate that training the network on additional data
with synthesized errors can improve performance.
\end{abstract}

\section{Introduction}


Systems that provide writing feedback
have great potential to assist language learners as well as native writers.
Although tools such as spell checkers have been useful,
detecting and fixing
errors in natural language, even at the sentence level,
remains far from solved.

Much of the prior research focuses solely on training classifiers for a small number
of error types, such as article or preposition errors \cite{han2006detecting,rozovskaya2010}.
More recent methods that consider a broader range of error classes
often rely on
language models to score $n$-grams or
statistical machine translation approaches \cite{ng2014conll}.
These methods, however, do not flexibly handle orthographic errors
in spelling, capitalization, and punctuation.

\begin{figure}[!t]
  \centering
      \includegraphics[width=0.40\textwidth]{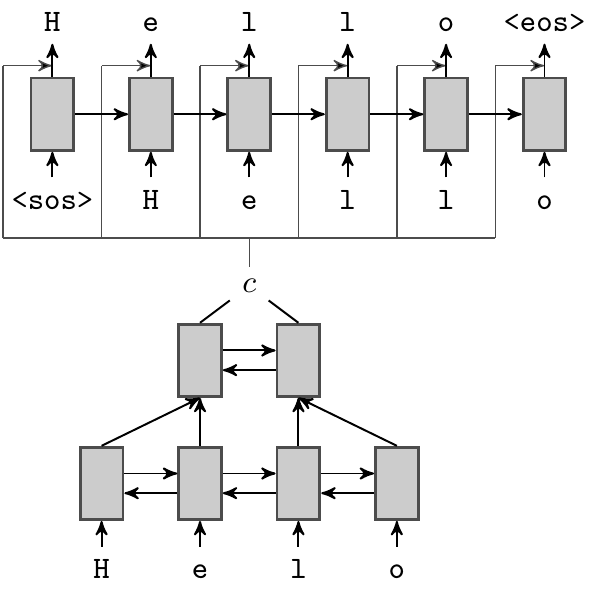}
  \caption{Illustration of the encoder-decoder neural network model with two encoder hidden layers and one decoder hidden layer. Character-level reasoning allows
  handling of misspellings and OOVs.}
  \label{fig:model_overview}
\end{figure}

As a motivating example, consider the following incorrect sentence:
``{\it I visitted Tokyo on Nov 2003. :)}''.
Several errors in this sentence illustrate the difficulties in the language correction setting.
First, the sentence contains a misspelling, {\it visitted}, an issue
for systems with fixed vocabularies. Second, the sentence contains rare words
such as {\it 2003} as well as punctuation forming an emoticon {\it :)}, issues
that may require special handling. Finally, the use of the
preposition {\it on} instead of {\it in} when not referring to a specific day is
non-idiomatic, demonstrating the complex patterns that must be captured to
suggest good corrections. In hopes of capturing such complex phenomena, we use
a neural network-based method.

Building on recent work in language modeling and machine translation,
we propose an approach to natural language error correction based on
an encoder-decoder recurrent neural network trained on a parallel corpus
containing ``good'' and ``bad'' sentences (Figure~\ref{fig:model_overview}).
When combined with
a language model, 
our system obtains state-of-the-art results on the CoNLL 2014 Shared Task,
beating systems using statistical machine translation systems, rule-based
methods, and task-specific features.
Our system naturally handles orthographic errors and rare words,
and can flexibly correct a variety of error types.
We further find that augmenting the network training data
with sentences containing synthesized errors
can result in significant gains in performance.

\section{Model Architecture}

Given an input sentence $x$ that we wish to map
to an output sentence $y$, we seek to model
$P(y | x)$.
Our model consists of an encoder and a decoder \cite{Sutskever2014,cho2014learning}.
The encoder maps the input sentence to a higher-level representation
with a pyramidal bi-directional RNN architecture similar to that
of \newcite{chan2015listen}.
The decoder is also a recurrent neural network that uses a content-based
attention mechanism \cite{bahdanau2014neural} to attend to the encoded representation and generate
the output sentence one character at a time.

\subsection{Character-Level Reasoning}
\label{ssec:character}

Our neural network model operates at the character level, in the encoder as
well as the decoder. This is for two reasons, as illustrated by our motivating example.
First, we do not assume that the inputs are spell-checked and often find spelling
errors in the sentences written by English learners in the datasets we consider.
Second, word-level neural MT models with a fixed vocabulary
are poorly suited to handle OOVs such as multi-digit numbers, emoticons, and
web addresses \cite{graves2013generating}, though recent work has proposed
workarounds for this problem \cite{luong2014addressing}.
Despite longer sequences in the character-based model,
optimization does not seem to be a significant issue, since
the network often only needs to copy characters from source to target.

\subsection{Encoder Network}
Given the input vector $x_t$,
the forward, backward, and combined activations of the
$j$th hidden layer are computed as:
\begin{align*}
f^{(j)}_t &= \mathrm{GRU}({f}^{(j)}_{t-1}, {c}^{(j-1)}_t),\\
b^{(j)}_t &= \mathrm{GRU}({b}^{(j)}_{t+1}, {c}^{(j-1)}_t),\\
h^{(j)}_t &= {f}^{(j)}_t + {b}^{(j)}_t
\end{align*}
where $\mathrm{GRU}$ denotes the gated recurrent unit function, which, similar to long short-term memory units (LSTMs), have shown to improve the performance of RNNs \cite{cho2014learning,hochreiter1997long}.

The input from the previous layer input ${c}^{(0)}_t = {x}_t$ and
\begin{equation*}
   {c}^{(j)}_t = \tanh\left({W}_\mathrm{pyr}^{(j)} \left[{h}^{(j-1)}_{2t}, {h}^{(j-1)}_{2t+1}\right]^\top + {b}_\mathrm{pyr}^{(j)}\right)
\end{equation*}
for $j>0$. The weight matrix ${W}_\mathrm{pyr}$ thus reduces the number of hidden
states for each additional hidden layer by half, and hence the encoder has a pyramid structure.
At the final hidden layer
we obtain the encoded representation ${c}$ consisting of
$\left \lceil T/2^{N-1} \right \rceil$ hidden states, where $N$
denotes the number of hidden layers.\vspace{0.3em}

\subsection{Decoder Network}
The decoder network is recurrent neural network using gated recurrent units
with $M$ hidden layers.
After the final hidden layer the network also conditions
on the encoded representation $c$ using an attention mechanism.

At the $j$th decoder layer the hidden activations are computed as
\begin{equation*}
d^{(j)}_t = \mathrm{GRU}(d^{(j)}_{t-1}, {d}^{(j-1)}_t),
\end{equation*}
with the output of the final hidden layer $d^{(M)}_t$ then being used as part of the content-based attention mechanism similar to that proposed by \newcite{bahdanau2014neural}:
\begin{align*}
    u_{tk} &= \phi_1(d^{(M)}_t)^\top\phi_2(c_k)\\
    \alpha_{tk} &= \frac{u_{tk}}{\sum_j u_{tj}}\\
    a_t &= \sum_j \alpha_{tj} c_j
\end{align*}
where $\phi_1$ and $\phi_2$ represent feedforward affine transforms followed by a $\tanh$ nonlinearity. The weighted sum of the encoded hidden states $a_t$ is then concatenated with $d^{(M)}_t$, and passed through another affine transform followed by a $\mathrm{ReLU}$ nonlinearity before the final softmax output layer.

The loss function is the cross-entropy loss per time step summed over the output sequence $y$:
\begin{equation*}
    L(x, y) = -\sum_{t=1}^T \log P(y_t | x, y_{<t}).
\end{equation*}
Note that during training the ground truth $y_{t-1}$ is fed into the network to predict $y_t$, while at test time the most probable $\hat{y}_{t-1}$ is used.
Figure~\ref{fig:model_overview} illustrates the
model architecture.

\subsection{Attention and Pyramid Structure}
\label{ssec:attention}

In preliminary experiments, we found that having an attention mechanism was crucial for the model
to be able to generate outputs character-by-character that did not diverge from the input.
While character-based approaches have not attained state-of-the-art performance on
large scale translation and language modeling tasks, in this setting the decoder network
simply needs to copy input tokens during the majority of time steps.

Although character-level models reduce the softmax over the vocabulary at each time step over word-level models, they also increase the total number of time-steps of the RNN. The content-based attention mechanism must then consider all the encoder hidden states $c_{1:T}$ at every step of the decoder.
Thus we use a pyramid architecture, which reduces computational complexity (as shown by \newcite{chan2015listen}).
For longer batches, we observe over a $2\times$ speedup for the same number of parameters when using a 400 hidden unit per layer
model with 3 hidden layers ($4\times$ reduction of steps in $c$).

\section{Decoding}

While it is simpler to integrate a language model
by using it as a re-ranker, here the language model probabilities
are combined with the encoder-decoder network through beam search.
This is possible because the attention mechanism in the decoder network
prevents the decoded output from straying too far from the source sentence.

\subsection{Language Model}

To model the distribution
\begin{equation}
P_\mathrm{LM}({y}_{1:T}) = \prod_{t=1}^T P({y}_t | {y}_{< t})
\end{equation}
we build a Kneser-Ney smoothed 5-gram language model on a subset of the
Common Crawl Repository\footnote{\tt http://commoncrawl.org} collected
during 2012 and 2013. After pruning, we obtain 2.2 billion $n$-grams.
To build and query the model, we use the KenLM toolkit \cite{Heafield2013}.

\subsection{Beam Search}

For inference we use a beam search decoder combining the neural network
and the language model likelihood. Similar to \newcite{hannun2014deep}, at step $k$, we rank the hypotheses
on the beam using the score
\begin{align*}
    s_k({y}_{1:k}|{x}) = \log P_\mathrm{NN}({y}_{1:k} | {x})
    + \lambda \log P_\mathrm{LM}({y}_{1:k})
\end{align*}
where the hyper-parameter $\lambda$ determines how much the language model is
weighted. To avoid penalizing longer hypotheses, we additionally normalize scores
by the number of words in the hypothesis $|{y}|$.
Since decoding is done at the character level,
the language model probability $P_\mathrm{LM}(\cdot)$ is only incorporated
after a space or end-of-sentence symbol is encountered.

\subsection{Controlling Precision}
\label{sec:edit_class}

For many error correction tasks, precision is emphasized more than recall;
for users, an incorrect suggestion is worse than a missed mistake.


In order to filter spurious edits, we train an edit classifier to classify
edits as correct or not. We run our decoder on
uncorrected sentences from our training data to generate
candidate corrected sentences. We then align the candidate sentences to
the uncorrected sentences by minimizing the word-level Levenshtein distance between
each candidate and uncorrected sentence. Contiguous segments that do not match
are extracted as proposed edits\footnote{Note this is an approximation and cannot distinguish side-by-side edits as separate edits.}.
We repeat this alignment and edit extraction process for the gold corrected sentences and the uncorrected sentences to obtain the gold edits.
``Good'' edits are defined as the intersection of the proposed and gold edits and ``bad'' edits are defined as the proposed edits not contained in the gold edits.
We compute edit features
and train a multilayer perceptron binary classifier
on the extracted edits to predict the probability of an edit being correct.
The features computed on an edit $s \rightarrow t$ are:
\begin{itemize}
    \item \textbf{edit distance features}: normalized word and character lengths of $s$ and $t$,
        normalized word and character insertions, deletions, and substitutions between $s$ and $t$.
    \item \textbf{embedding features}: sum of 100 dimensional GloVe \cite{pennington2014glove} vectors of words in $s$ and $t$, GloVe vectors of left and right context words in $s$.
\end{itemize}


In order to filter incorrect edits, we only accept edits whose predicted
probability exceeds a threshold $p_\mathrm{min}$.
This assumes that classifier probabilities are reasonably calibrated
\cite{niculescu2005predicting}.
Edit classification improves precision with a small drop in recall;
most importantly, it helps filter edits where the decoder network
misbehaves and $t$ deviates wildly from $s$.

\section{Experiments}

We perform experiments using two datasets of corrected sentences written
by English learners. The first is the Lang-8 Corpus,
which contains erroneous sentences and their corrected versions
collected from a social language learner forum \cite{tajiri2012tense}.
Due to the online user-generated setting, the Lang-8 data is
noisy, with sentences often containing misspellings, emoticons,
and other loose punctuation. Sample sentences are show in Table~\ref{tab:good_samples}.

The other dataset we consider comes from the CoNLL 2013 and 2014 Shared Tasks,
which contain about 60K sentences from essays written by English learners with
corrections and error type annotations. We use the larger Lang-8 Corpus
primarily to train our network, then evaluate on the CoNLL Shared Tasks.

\subsection{Training and Decoding Details}

Our pyramidal encoder has $3$ layers, resulting in a factor $4$
reduction in the sequence length at its output, and our decoder RNN
has $3$ layers as well. Both the encoder and decoder use a hidden size of $400$ and gated recurrent units (GRUs),
which along with LSTMs \cite{hochreiter1997long} have been shown to
be easier to optimize and preserve information over many time
steps better than vanilla recurrent networks.

Our vocabulary includes 98 characters: the printing ASCII character set and
special
$\langle$sos$\rangle$,
$\langle$eos$\rangle$,
and $\langle$unk$\rangle$\ symbols indicating the start-of-sentence, end-of-sentence, and unknown symbols, respectively.

To train the encoder-decoder network we use the Adam optimizer \cite{kingma2014adam}
with a learning rate of $0.0003$, default decay rates $\beta_1$ and $\beta_2$, and
a minibatch size of 128.
We train for up to $40$ epochs, selecting the model with the lowest perplexity on the Lang-8 development set.
We found that using dropout \cite{srivastava2014dropout} at a rate of $0.15$ on the non-recurrent connections \cite{pham2014dropout} helped reduce perplexity.
We use uniform initialization of the weight matrices in the range $[-0.1, 0.1]$ and zero initialization of biases.

Decoding parameter $\lambda$ and edit classifier threshold $p_\mathrm{min}$
were chosen to maximize
performance on the development sets of the datasets described.
All results were obtained using a beam width of 64, which seemed to provide
a good trade-off between speed and performance.

\subsection{Noisy Data: Lang-8 Corpus}

\begin{table}[bt]
\begin{center}
\begin{tabular}{l  r}
\toprule
Method & Test BLEU \\
\midrule
No edits & 59.54 \\
Spell check & 58.91 \\
RNN &  61.63 \\
RNN + LM & \textbf{61.70} \\
\bottomrule
\end{tabular}
\end{center}
\caption{Performance on Lang-8 test set. Adding the language model
results in a negligible increase in performance, illustrating the difficulty
of the user-generated forum setting.}
\label{tab:lang8}
\end{table}

We use the train-test split provided by the Lang-8 Corpus of Learner English \cite{tajiri2012tense},
which contains 100K and 1K entries with about 550K and 5K parallel sentences, respectively.
We also split 5K sentences from the training set to use as a separate development set
for model and parameter selection.

Since we do not have gold annotations that distinguish side-by-side edits as separate edits,
we report BLEU score\footnote{Using case-sensitive \texttt{multi-bleu.perl} from Moses.}
using just the encoder-decoder network as well as when combined with the $n$-gram language model
(Table~\ref{tab:lang8}). Note that since there may be multiple ways to correct an error
and some errors are left uncorrected, the baseline of using uncorrected sentences is more difficult
to improve upon than it may initially appear. As another baseline we apply the top
suggestions from a spell checker with default configurations\footnote{Hunspell v1.3.4, {\tt \url{https://hunspell.github.io}}}.
We suspect due to proper nouns, acronyms, and inconsistent capitalization conventions
in Lang-8, however, this actually decreased BLEU slightly. To the
best of our knowledge, no other work has reported results on this challenging task.

\subsection{Main Results: CoNLL Shared Tasks}

\begin{table}[bt]
\begin{center}
\begin{tabular}{l r r r}
\toprule
Method & $P$ & $R$ & $F_{0.5}$\\
\midrule
RNN & 42.96 & 6.27 & 19.81 \\
RNN aug & 49.30 & 10.10 & 27.75 \\
RNN + LM & 43.27 & 15.14 & 31.55 \\
RNN aug + LM & 46.94 & \textbf{17.11} & 34.81 \\
RNN aug + LM + EC & \textbf{51.38} & 15.83 & \textbf{35.45}\\
\bottomrule
\end{tabular}
\end{center}
\caption{Development set performance. EC denotes edit classification (Section~\ref{sec:edit_class}), and ``aug'' indicates data augmentation was used.}
\label{tab:conll2013}
\end{table}


\begin{table}[bt]
\begin{center}
\begin{tabular}{l  r  r  r }
\toprule
Method &  $P$ & $R$ & $F_{0.5}$ \\
\midrule
AMU & 41.62 & 21.40 & 35.01 \\
CUUI & 41.78 & 24.88 & 36.79 \\
CAMB & 39.71 & 30.10 & 37.33 \\
\newcite{susanto2015systems} & 53.55 & 19.14 & 39.39 \\
Ours (no EC) & 45.86 & 26.40 & 39.97 \\
Ours (+ EC) & 49.24 & 23.77 & \textbf{40.56}\\
\midrule
Ours (A1) & 32.56 & 14.76 & 26.23 \\
Ours (A2) & 44.04 & 14.83 & 31.59 \\
A1 (A2) & 50.47 & 32.29 & \textbf{45.36} \\
A2 (A1) & 37.14 & 45.38 & \textbf{38.54} \\
\bottomrule
\end{tabular}
\end{center}
\caption{CoNLL 2014 test set performance. We compare to the 3 best CoNLL 2014
submissions which used combinations of MT, LM ranking, and error type-specific classifiers.
We report $F$-score against both and single annotators,
as well as each annotator scored against the other as a human ceiling.
A1 and A2 denote Annotators 1 and 2.}
\label{tab:conll2014}
\end{table}

\sssection{Description}
For our second set of experiments we evaluate on the CoNLL 2014 Shared Task on Grammatical Error Correction \cite{ng2013conll,ng2014conll}.
We use the revised CoNLL 2013 test data with all error types as a development
set for parameter and model selection with the 2014 test data as our test set.
The 2013 test data contains 1381 sentences with 3470 errors in total,
and the 2014 test data contains 1312 sentences with 3331 errors.
The CoNLL 2014 training set contains 57K sentences with the corresponding gold edits by a single
annotator. The 2013 test set is also only labeled by a single annotator, while the 2014 test set
has two separate annotators.

We use the NUS MaxMatch scorer \cite{dahlmeier2012better} v3.2 in order to compute the
precision ($P$), recall ($R$), and $F$-score for our corrected sentences. Since precision is considered more
important than recall for the error correction task, $F_{0.5}$ score is reported as in the CoNLL 2014 Challenge.
We compare to the top submissions in the 2014 Challenge as well as the method by \newcite{susanto2015systems},
which combines 3 of the weaker systems to achieve the state-of-the-art result.
All results reported on the 2014 test set exclude alternative corrections submitted by the participants.

\sssection{Synthesizing Errors}
In addition to the Lang-8 training data, we include the CoNLL 2014
training data in order to train the encoder-decoder network.
Following prior work, we additionally explore synthesizing
additional sentences containing errors using the CoNLL 2014 training data
\cite{felice2014generating,rozovskaya2012ui}.
Our data augmentation procedure generates synthetic errors
for two of the most common error types in the development set:
article or determiner errors (ArtOrDet) and noun number errors (Nn).
Similar to \newcite{felice2014generating},
we first collect error distribution statistics
from the CoNLL 2014 training data. For ArtOrDet errors, we estimate
the probability that an article or determiner
is deleted, replaced with another determiner, or inserted before the start
of a noun phrase. For Nn errors, we
estimate the probability that it is replaced with its singular or plural form.
To obtain sentence parses we use the Stanford CoreNLP Toolkit \cite{manning2014}.
Example synthesized errors:
\begin{itemize}
    \item \textbf{ArtOrDet}: They will generate and brainstorm \underline{\it the} innovative ideas.
    \item \textbf{Nn}: Identification is becoming more important in our {\it society} $\rightarrow$ \underline{\it societies}.
\end{itemize}
Errors are introduced independently
according to their estimated probabilities
by iterating over the words in the training sentences,
and we produce two corrupted versions of each training sentence
whenever possible.
    The original Lang-8 training data contains 550K sentence pairs. Adding the CoNLL 2014 training data results in
    about 610K sentence pairs, and after data augmentation we obtain a total of 720K sentence pairs. We examine the benefits of synthesizing errors in Section~\ref{sec:discussion}.

\sssection{Results} Results for the development set are shown in Table~\ref{tab:conll2013},
and results for the CoNLL 2014 test set in Table~\ref{tab:conll2014}.
On the CoNLL 2014 test set, which contains the full set of 28 error types,
our method achieves a state-of-the-art result, beating all systems from the 2014 Challenge
as well as a system combination method \cite{susanto2015systems}.
Methods from the 2014 Challenge used statistical machine translation, language model ranking,
rule-based approaches, and error type-specific features and classifiers, often in combination.
System descriptions for participating teams are given in \newcite{ng2014conll}.

\section{Discussion}
\label{sec:discussion}

\begin{table*}[bt]
    \newcounter{rownum}
    \newcommand\rownumber{\stepcounter{rownum}\textcolor{gray}{\arabic{rownum}}}
\begin{center}
    \begin{tabular}{r p{0.45\textwidth} p{0.45\textwidth}}
\toprule
& {\normalsize Original} & {\normalsize Proposed}\\
\midrule
\rownumber & It 's \underline{\it heavy rain} today & It 's \underline{\it raining heavily} today\\
\rownumber & Everyone wants to be \underline{\it success} . & Everyone wants to be \underline{\it successful} .\\
\rownumber & \underline{\it On the} 3 weeks , I learned many things . & \underline{\it In the last} 3 weeks , I learned many things . \\
\rownumber & \underline{\it this is the} first entry ! : D & \underline{\it This is my} first entry ! : D\\
\rownumber & Help me \underline{\it getting English skill} , please . & Help me \underline{\it improve my English skills} , please .\\
\rownumber & \underline{\it At} last night , the 24th of June 2010 was big night for the Japanese national team and heaps of fans . & Last night , the 24th of June 2010 was \underline{\it a} big night for the Japanese national team and heaps of fans .\\
\rownumber & I \underline{\it start to learning} English again . & I \underline{\it am starting to learn} English again .\\
\rownumber & I went to \underline{\it Beijin} in China \underline{\it for} four days \underline{\it in} this week . & I went to \underline{\it Beijing} in China four days this week .\\
\rownumber & After a long day , \underline{\it I and my roommate} usually sit down , drink coffee and listen to music . & After a long day , \underline{\it my roommate and I} usually sit down , drink coffee and listen to music .\\
\rownumber & Do you know \underline{\it a Toeic} ? & Do you know about \underline{\it TOEIC} ?\\
\toprule
\end{tabular}
\end{center}
\caption{Sample edits from our Lang-8 development set using only the character-based encoder-decoder network. Note that the model is able to handle misspellings (\textit{Beijin}) as well as rare words (\textit{TOEIC}) and emoticons (\it{:~D}).}
\label{tab:good_samples}
\end{table*}

\begin{table*}[bt]
    \newcounter{rownum2}
    \newcommand\rownumber{\stepcounter{rownum2}\textcolor{gray}{\arabic{rownum2}}}
\begin{center}
    \begin{tabular}{l p{0.45\textwidth} p{0.45\textwidth}}
\toprule
& {\normalsize Original} & {\normalsize Proposed}\\
\midrule
\rownumber & Broke my heart & \underline{\it I} broke my heart\\
\rownumber & I want to big size bag & I want to \underline{\it be} a big size bag\\
\rownumber & This is typical Japanese male \underline{\it hobit} & This is \underline{\it a} typical Japanese male \underline{\it hobby}\\
\rownumber & I 'm so sorry to miss Lukas \underline{\it Moodysson 's} Lijia 4-ever . & I 'm so sorry to miss Lukas \underline{\it Moodysnot} Lijia 4-ever .\\
\rownumber & The match is the Rockets \underline{\it withthe} Bulls . & The match is the Rockets \underline{\it withth} Bulls .\\
\bottomrule
\end{tabular}
\end{center}
\caption{Sample incorrect and ambiguous edits, again using just the encoder-decoder network.}
\label{tab:bad_samples}
\end{table*}

\sssection{Qualitative Analysis}
We present examples of correct and incorrect edits on Lang-8 development set in Table~\ref{tab:good_samples} and Table~\ref{tab:bad_samples}.
Despite operating at the character level, the network is occasionally able to perform rearrangements of words to form
common phrases (e.g. {\it I and my roommate} $\rightarrow$ {\it my roommate and I}) and insert and delete words where appropriate.
On the other hand, the network can also sometimes mangle rare words ({\it Moodysson} $\rightarrow$ {\it Moodysnot})
and fail to split common words missing a separating space ({\it withthe} $\rightarrow$ {\it withth}), suggesting that while common patterns are captured, the
network lacks semantic understanding.

\begin{figure}[!ht]
  \centering
      \includegraphics[width=0.4\textwidth]{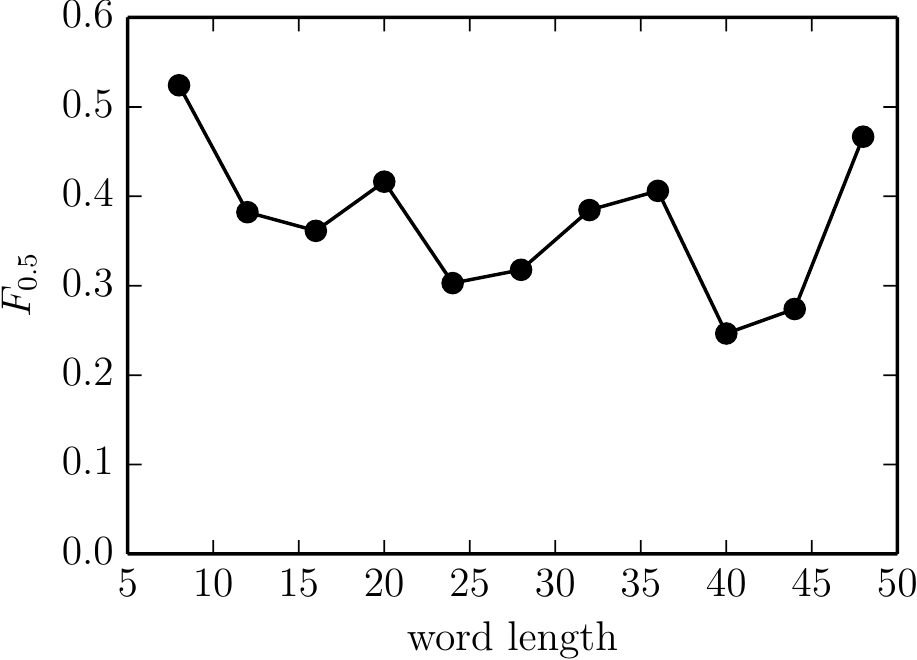}
  \caption{$F$-score vs. length of input sentence on development set. Only bins with $10$ or more sentences included.}
  \label{fig:perf_length}
\end{figure}

\sssection{Performance Breakdown}
While the encoder-decoder network can itself produce modifications,
on less noisy datasets such as the CoNLL Challenge datasets a language model
can greatly improve performance.
Increasing the language model weight $\lambda$ tends to
improves recall at the expense of precision. 
On the other hand, using edit classification to filter spurious edits
increases precision, often with smaller drops in recall.
We do not observe a trend of decreasing $F$-score for a wide range of sentence lengths (Figure~\ref{fig:perf_length}), likely
due to the attention mechanism, which
helps to prevent the decoded output from diverging from the input sentence.

\begin{table}[bt]
\begin{center}
\begin{tabular}{l  r  r  r }
\toprule
Type & Count & $R$ no aug & $R$ aug \\
\midrule
\textbf{ArtOrDet} & 717 & 20.08 & 29.14 \\
Wci & 434 & 2.30 & 1.61 \\
\textbf{Nn} & 400 & 31.50 & 51.00 \\
Preposition & 315 & 13.01 & 7.93 \\
Word form & 223 & 26.90 & 19.73 \\
\bottomrule
\end{tabular}
\end{center}
\caption{CoNLL development set recall for 5 most frequent error categories with and without training on data with synthesized article/determiner and noun number errors. Wci denotes wrong collocation/idiom errors.}
\label{tab:aug_err_types}
\end{table}

We report the inter-annotator agreement in Table~\ref{tab:conll2014},
which gives a possible bound on the $F$-score for this task.

\sssection{Effects of Data Augmentation}
We obtain promising improvements using data augmentation,
boosting $F_{0.5}$-score on the development set from 31.55 to 34.81.
For the two error types where we synthesize data (article or determiner and noun number)
we observe significant increases in recall, as shown in Table~\ref{tab:aug_err_types}. The same phenomenon has been observed by \newcite{rozovskaya2012ui}.
Interestingly, the recall of other error types (see \newcite{ng2014conll} for descriptions) decreases.
We surmise this is because the additional training data contains
only ArtOrDet and Nn errors, and hence the network is encouraged to simply
copy the output when those error types are not present.
We hope synthesizing data with a variety of other error types may fix this
issue and improve performance.

\begin{table*}[bt]
\begin{center}
\begin{tabular}{l  l  l  l }
\toprule
Type & Description & Original & Proposed \\
\midrule
Mec & Spelling, punctuation, & Another identification is  & Another identification is \\
& capitalization, etc.   & \underline{\it Implanting} RFID chips ... & \underline{\it implanting} RFID chips ... \\
Rloc- & Redundancy & \dots it seems that our freedom \underline{\it of} & \dots it seems that our freedom \\
&              &  \underline{\it doing things} is being invaded.               & is being invaded. \\
             Wci & Wrong collocation/idiom & Every coin has \underline{\it its} two sides. & Every coin has two sides. \\
\bottomrule
\end{tabular}
\end{center}
\caption{Examples of the aforementioned challenging error types that our system fixed.}
\label{tab:challenging_err_types}
\end{table*}
\vspace{1em} 

\sssection{Challenging Error Types}
We now examine a few illustrative error types
from the CoNLL Challenges that originally motivated our approach:
orthographic (Mec),
redundancy (Rloc-), and idiomatic errors (Wci).
Since the 2013 Challenge did not score these error types,
we compare our recall to those of participants in the 2014 Challenge \cite{ng2014conll}.\footnote{The team that placed 9th overall did not disclose their method; thus we only compare to the 12 remaining teams.} Note that systems
only predict corrected sentences and not error types, and hence precision is not compared.
We use the results from our final system, including both data augmentation
and edit classification. Some examples of these error types are shown in Table~\ref{tab:challenging_err_types}.

\begin{itemize}
    \item \textbf{Mec}: We obtain a recall of 37.17 on the Mec error type,
          higher than all the 2014 Challenge teams besides
          one team (RAC) that used rule-based methods to attain 43.51 recall.
          The word/phrase-based translation and language modeling approaches do not
          seem to perform as well for fixing orthographic errors.
    \item \textbf{Rloc-}: Redundancy is difficult to capture using just rule-based
        approaches and classifiers; our approach obtains 17.47 recall which
        places second among the 12 teams. The top system
        obtains 20.16 recall using a combination MT, LM, and rule-based method.
    \item \textbf{Wci}: Although there are 340 collocation errors,
        all teams performed poorly on this category. Our recall places 3rd
        behind two teams (AMU and CAMB) whose methods both used an MT system. Again,
        this demonstrates the difficulty of capturing whether a
        sentence is idiomatic through
        only classifiers and rule-based methods.
\end{itemize}
We note that our system obtains significantly higher precision
than any of the top 10 teams in the 2014 Challenge (49.24 vs. 41.78),
which comes at the expense of recall.

\sssection{Limitations}
A key limitation of our method as well as most other
translation-based methods is that it is trained on just parallel sentences,
despite some errors requiring information about
the surrounding text to make the proper correction.
Even within individual sentences, when longer context is needed to
make a correction (for example in many subject-verb agreement errors), the performance is hit-and-miss.
The edits introduced by the system tend to be fairly local.

Other errors illustrate the need for natural language understanding,
for example in Table~\ref{tab:bad_samples} the correction \textit{Broke my heart}
$\rightarrow$ \textit{\underline{I} broke my heart} and \textit{I want to big size bag} $\rightarrow$ \textit{I want to \underline{be} a big size bag}.
Finally, although end-to-end approaches have the potential to fix a
wide variety of errors, it is not straightforward to then classify the types of errors
being made. Thus the system cannot easily provide error-specific feedback.

\section{Related Work}

Our work primarily builds on prior work on training encoder-decoder RNNs for machine
translation \cite{kalchbrenner2013recurrent,Sutskever2014,cho2014learning}.
The attention mechanism, which allows the decoder network to copy parts of the source
sentence and cope with long inputs, is based on the content-based attention mechanism
introduced by \newcite{bahdanau2014neural}, and the overall network architecture
is based on that described by \newcite{chan2015listen}.
Our model is also inspired by character-level models as proposed by
\newcite{graves2013generating}. More recent work has applied character-level models
to machine translation and speech recognition as well, suggesting
that it may be applicable to many other tasks that involve the
problem of OOVs
\cite{ling2015character,lexfree2015,chan2015listen}.

Treating grammatical error correction as a statistical machine translation
problem is an old idea; the method of mapping ``bad'' to ``good'' sentences was
used by many of the teams in the CoNLL 2014 Challenge \cite{Felice14grammaticalerror,junczys2014}.
The work of \newcite{Felice14grammaticalerror} achieved the best $F_{0.5}$-score of 37.33 in
that year's challenge using a combination of rule-based, language-model ranking, and statistical machine translation techniques. Many other teams used a
language model for re-ranking hypotheses as well.
Other teams participating in the CoNLL 2014 Challenge used techniques ranging from
rule-based systems to type-specific classifiers, as well as combinations of the two
\cite{rozovskaya-EtAl:2014:W14-17,lee-lee:2014:W14-17}. The rule-based systems
often focus on only a subset of the error types.
The previous state of the art was achieved by \newcite{susanto2015systems}
using the system combination method proposed
by \newcite{Heafield2010} to combine three weaker systems.

Finally, our work uses data collected and shared through the generous efforts of the teams behind the CoNLL and Lang-8 datasets
\cite{mizumoto2011mining,hayashibe2012effect,ng2013conll,ng2014conll}.
Prior work has also proposed data augmentation for the language correction task
\cite{felice2014generating,rozovskaya2012ui}.

\section*{Conclusion}

We present a neural network-based model for performing language correction.
Our system is able correct
errors on noisy data collected from an English learner forum and attains state-of-the-art
performance on the CoNLL 2014 Challenge dataset of annotated essays.
Key to our approach is the use of a character-based model with
an attention mechanism, which allows for orthographic errors to be captured
and avoids the OOV problem suffered by word-based neural machine translation methods.
We hope the generality of this approach will also allow it to be
applied to other tasks that must deal with noisy text,
such as in the online user-generated setting.

\section*{Acknowledgments}


We thank Kenneth Heafield, Jiwei Li,
Thang Luong, Peng Qi, and Anshul Samar for helpful discussions.
We additionally thank the developers of Theano \cite{bergstra2010theano}.
Some GPUs used in this work were donated by NVIDIA Corporation.
ZX was supported by an NDSEG Fellowship.
This project was funded in part by DARPA MUSE award FA8750-15-C-0242 AFRL/RIKF.

\bibliography{dl}
\bibliographystyle{acl2016}

\end{document}